%% file: sokoban_tmlr.tex
\documentclass{article}
\usepackage[preprint]{tmlr}
\usepackage[utf8]{inputenc}
\usepackage{graphicx}
\usepackage{footmisc}
\usepackage{url}
\usepackage{amsmath}
\usepackage[urlcolor=blue,citecolor=blue]{hyperref}
\usepackage{caption}
\usepackage{subcaption}
\usepackage[most]{tcolorbox} \tcbuselibrary{listingsutf8}
\usepackage{listings}
\usepackage{listingsutf8}
\usepackage{multirow}
\usepackage{booktabs}
\input{math_commands}

\usepackage{textcomp}
\usepackage{amsmath,amsfonts}
\usepackage{graphicx}
\usepackage{caption}
\usepackage{booktabs}
\usepackage{multirow}
\usepackage{xcolor}
\usepackage{array}
\usepackage{float}

\title{SokoBench: Evaluating Long-Horizon Planning and Reasoning in Large Language Models}

\author{
\name Sebastiano Monti \email s.monti@ipazia.com \addr Ipazia SpA, Italy
\AND
\name Carlo Nicolini \email c.nicolini@ipazia.com \addr Ipazia SpA, Italy
\AND
\name Giovanni Pellegrini \email g.pellegrini@ipazia.com \addr Ipazia SpA, Italy
\AND
\name Jacopo Staiano \email jacopo.staiano@unitn.it \addr University of Trento, Italy
\AND
\name Bruno Lepri \email lepri@fbk.eu \addr Fondazione Bruno Kessler and Ipazia SpA, Italy
}

\begin{document}
\maketitle

\begin{abstract}
Although the capabilities of large language models have been increasingly tested on complex reasoning tasks, their long-horizon planning abilities have not yet been extensively investigated. In this work, we provide a systematic assessment of the planning and long-horizon reasoning capabilities of state-of-the-art Large Reasoning Models (LRMs). We propose a novel benchmark based on Sokoban puzzles, intentionally simplified to isolate long-horizon planning from state persistence. Our findings reveal a consistent degradation in planning performance when more than 25 moves are required to reach the solution, suggesting a fundamental constraint on forward planning capacity. We show that equipping LRMs with Planning Domain Definition Language (PDDL) parsing, validation, and solving tools allows for modest improvements, suggesting inherent architectural limitations which might not be overcome by test-time scaling approaches alone. 
\end{abstract}

\section{Introduction}

Automated Planning, i.e. the task of generating sequences of actions to achieve a goal, is a well-studied problem in the field of Artificial Intelligence (AI) \citep{ghallab2016automated}, since it requires AI systems to exhibit cognitive abilities such as reasoning, understanding, and efficient state space search \citep{wei2025plangenllms}. 
To this end, automated planning literature has focused on the use of formal languages, such as the Planning Domain Definition Language (PDDL) \citep{mcdermott1998,russell2021artificial,haslum2019introduction}), and of tree-search strategies or specific heuristics to find optimal solutions \citep{bonet2001planning}.
Large Language Models (LLMs) and, in particular, Large Reasoning Models (LRMs) i.e., LLMs trained to produce so-called \emph{reasoning traces} resembling structured thought processes, have demonstrated impressive capabilities in natural language understanding, knowledge retrieval and multi-modal pattern recognition \citep{jaech2024openai,guo2025deepseek,team2025kimi}.
However, recent studies highlighted the limitations of such models when applied to planning tasks \citep{valmeekam2023planning,shojaee2025illusion}.
For instance, internal reasoning processes have been shown to resemble a form of wandering through the solution space rather than a systematic exploration \citep{lu2025reasoning}. 
This distinction becomes particularly important for problems that require maintaining sequential state information, such as spatial exploration in constrained environments. In these settings, effective tracking of working memory is necessary to infer the agent’s latent previous state \citep{zhang2024working}.

In this work, we investigate the long-horizon planning abilities of LRMs using a highly simplified variant of the Sokoban puzzle~\citep{culberson1997sokoban}. Rather than increasing spatial complexity, we \emph{deliberately minimize the structural complexity} of the environment while preserving the long-horizon nature of the task by creating examples with 
the lowest possible branching factor compatible with solvability: 
a single movable block placed within a linear corridor with tightly controlled geometry.

This setting allows us to isolate long-horizon planning from state persistence: models are required to produce complete solution sequences without external memory, intermediate feedback, or state validation, relying solely on internal state representations to track the evolving environment.
We therefore investigate to what extent LRMs can sustain coherent planning over long (but simple) action sequences and whether even minimal reasoning branching in otherwise trivial Sokoban instances is sufficient to induce planning failures.

Concretely, we examine whether current LRMs can reliably solve linear-corridor Sokoban puzzles with minimal possible branching and identify the point at which increases in horizon length lead to catastrophic breakdowns in action validity, despite the simplicity of the underlying environment.
As we will show these minimal sub-problems which are trivial to humans \citep{jarusek2010difficulty}, are still challenging for Large Reasoning Models as shown by other preliminary studies involving spatial intelligence \citep{cai2025spatialintelligence}.
We posit this as a systemic deficiency in long-term action representation and sequential logic, and in spatial reasoning and thus as an important limitation of current LRMs that is not yet fully understood.


\section{Related Work}
\subsection{Benchmarks for LLM Planning}
As mentioned above, planning requires LLMs to blend logical, numerical, and spatial reasoning with long-horizon strategic adaptation, rather than just relying on pattern matching or memorization.
Classical planning domains expressed in or derived from the Planning Domain Definition Language (PDDL \citep{fox2003pddl2}), such as BlocksWorld \citep{slaney2001blocks}, Towers of Hanoi and similar tasks  \citep{pallagani2023understanding}, remain a common benchmark choice, though earlier attempts date back to the pre-ChatGPT era \citep{silver2022pddl}.
Test suites like PlanBench \citep{planbench2022} introduced structured, domain-agnostic evaluations inspired by classical planning \citep{ghallab2016automated}, including plan generation \citep{oswald2024large,valmeekam2025a,lamalfa2025end} and optimality \citep{planbench2022,zhai2025llmplanbench,valmeekam2023planbench}.

In another line of work, planning is evaluated within \textit{agentic} or workflow-based frameworks, where LLMs are required to decompose goals into multiple sub-plans \citep{meyerson2025solving,zhang2025recap,lamalfa2025end}. 
The results in these settings are encouraging though highly cost intensive. 
Importantly, when not equipped with external tools or made part of larger workflows (e.g., enabling stateful tracking \citep{hu2025agentgen}), innate planning abilities remain still weak \citep{schepanowski2025limits}.
Even the latest foundational models are found to consistently fail in delivering correct sequences of actions (in any format or language) due to two primary deficits: weak internal state representations leading to invalid moves and misleading heuristic search resulting in loops or early termination, as shown in the textual game ``8-puzzle'' in \cite{schepanowski2025limits}.
Moreover, efficacy of different prompting techniques is model-dependent in a non-predictable way \citep{schepanowski2025limits,deng2025can}.

Other works have systematically investigated the performances of LLMs in playing textual games with \textit{gym}-style APIs \citep{brockman2016,hu2025lmgame}.
Beyond structured puzzles, community-driven and informal game-oriented benchmarks like word-game bench \citep{stojanovski2024} and nonogram logic puzzles \citep{berend2014nonograms,kleine2016} with multi-difficulty instances have been devised to measure how well models plan under both explicit and implicit constraints, track environment states, and adapt over multiple turns.
The varying depth of planning ability required helps to reveal how performance scales with complexity and structure.

In general, existing benchmarks using specific planning languages and/or internal reasoning traces expressed in natural language show that LLMs exhibit limited planning abilities in various domains \citep{kambhampati2024position}, especially as the complexity and horizon length of the problems increase.
This gap motivates the development of new benchmarks tailored to planning and solving structured textual puzzles with LLMs.

\subsection{Sokoban as a Benchmark for Planning} 
The Sokoban puzzle involves spatial planning in a highly constrained environment. Solvable Sokoban maps can be generated efficiently \citep{murase1996automatic}, and the environment is fully controllable and deterministic. These properties enable rigorous evaluation using exact solvers and verifiers, as well as metrics such as search depth and solution time \citep{jarusek2010difficulty, shoham2020fess}.
Unlike puzzles such as the Tower of Hanoi, which can be solved by repeating a simple pattern for larger instances, Sokoban offers no shortcuts. 
Each map is unique, and moving a single box can block or open paths in ways that prevent a one-size-fits-all solution.
As a result, Sokoban is considered a good benchmark for evaluating planning abilities in the 2023 edition of the International Planning Competition~\citep{taitler2023}. 

Recently, recurrent neural networks (non LLM-based) trained over multiple examples of Sokoban puzzles have obtained state of the art performance \citep{jolicoeur2025less,taufeeque2024planning}. 
However, LLMs are found to perform poorly, struggling even with simple maps and correctly solving only a small fraction of instances: \cite{valmeekam2025a} report success rates of just about 10–12\% when using the OpenAI \texttt{o1-preview} model directly. 
In contrast, substantially higher success rates are achieved in an LLM-Modulo setting, where the same model is used to generate plans that are then executed by an external planner, yielding approximately 43\% solved instances for \texttt{o1-preview} (and about 10\% for \texttt{o1-mini}), albeit at significantly higher computational cost.

Most prior work on textual puzzle solving and planning with LLMs has emphasized high-level notions such as search depth, branching factor, or overall puzzle complexity. 
Much less attention has been paid to the role of simpler, low-level operations that these tasks implicitly rely on. 
Evidence from seemingly trivial problems suggests that LLM failures do not always stem from complexity itself, but from how basic reasoning steps are elicited. 
A well-known example is the character-counting question “how many r’s are in strawberry?” \citep{karpathy2024tokenization}, which has sparked debate over whether LLM errors are caused by tokenization or deeper representational limits~\citep{shin2024large}. 
The work by \cite{xu2025llm} revisits this issue through a careful empirical study, showing that LLMs are in fact capable of performing these simple symbolic operations, but often fail unless prompted to reason explicitly. 
Character-level benchmarks, such as CharBench \citep{uzan2025charbench}, shows that modern LLMs struggle with simple character counting and positioning tasks not because tokenization fully explains these errors, but because intrinsic properties like word length and actual character count have a stronger influence on performance, indicating that basic symbolic operations are not reliably deployed unless the model is guided to engage them explicitly.

Put together, these observations point to a broader interpretation of failures in spatial planning and puzzle games, suggesting that they may arise from missing or weak activation of basic operations, rather than from the inherent difficulty of the planning problem.

\section{Methods}
\subsection{Sokoban game}
\begin{figure}
\centering
\begin{tabular}{@{}p{0.48\textwidth}@{\hspace{1em}}p{0.48\textwidth}@{}}
\begin{minipage}[ht]{\linewidth}
    \centering
    \includegraphics[width=\linewidth]{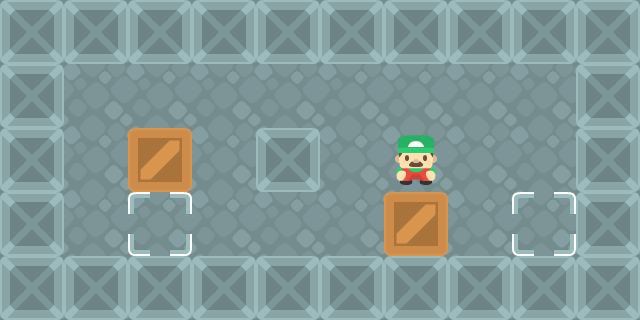}

    \captionof{figure}{Example of a Sokoban puzzle. All boxes must be pushed onto goal positions.
    A solution to this problem in compressed notation is:
    \textcolor{green}{\texttt{1}$\boldsymbol{\uparrow}$},
    \textcolor{red}{\texttt{4}$\boldsymbol{\leftarrow}$},
    \textcolor{blue}{\texttt{1}$\boldsymbol{\downarrow}$},
    \textcolor{orange}{\texttt{1}$\boldsymbol{\rightarrow}$},
    \textcolor{blue}{\texttt{1}$\boldsymbol{\downarrow}$},
    \textcolor{orange}{\texttt{4}$\boldsymbol{\rightarrow}$},
    resulting in the LURD notation
    \texttt{u,l,l,l,l,D,r,d,r,r,r,R}.}
    \label{fig:soko_2d}
\end{minipage}&
\begin{minipage}[ht]{\linewidth}
\centering
\textbf{Equivalent ASCII format}

{\ttfamily
\setlength{\tabcolsep}{4pt}
\renewcommand{\arraystretch}{1.1}
\begin{tabular}{cccccccccc}
\# & \# & \# & \# & \# & \# & \# & \# & \# & \# \\
\# &   &   &   &   &   &   &   &   & \# \\
\# &   & \$ &   & \# &   & @ &   &   & \# \\
\# &   & .  &   &   &   & \$ &   & . & \# \\
\# & \# & \# & \# & \# & \# & \# & \# & \# & \# \\
\end{tabular}
}

\small
\begin{tabular}{l|c|c}
\textbf{Game} & \textbf{Map Element} & \textbf{ASCII Symbol} \\
\midrule
\multirow{6}{*}{Sokoban}
& Player         & @ \\
& Player on Goal & + \\
& Box            & \$ \\
& Box on Goal    & * \\
& Goal           & . \\
& Wall Brick     & \# \\
\end{tabular}

\vspace{0.6em}

\captionsetup{width=0.9\linewidth}
\captionof{table}{ASCII notation of the elements of Sokoban maps. Empty areas are encoded as space (\textvisiblespace).}
\label{tab:ascii-sokoban-notation}

\end{minipage}

\end{tabular}
\end{figure}

Figure~\ref{fig:soko_2d} shows an example of a Sokoban puzzle and the game's central mechanic: the player controls a sprite that pushes boxes within a two-dimensional spatially constrained environment with the goal to position them onto predefined locations.
Despite its apparent simplicity, Sokoban is a NP-hard and PSPACE-complete problem \citep{culberson1997sokoban}, positioning it as a canonical domain for symbolic and hierarchical planning. 
Apart from the pictorial representation, Sokoban maps can be encoded using an ASCII-based symbolic representation as expressed in Table~\ref{tab:ascii-sokoban-notation}.
Sequences of main character actions are typically encoded in LURD format (left,up,right,down), with lowercase letters indicating simple moves, and uppercase letters indicating box pushes.
Although moves and pushes have distinct notations in classical Sokoban planning, in our experiments we restrict only to comma-separated uppercase letters.
This representation does not compromise the information content of the solutions and simplifies the output format for language models, avoiding potential mistakes.

\subsection{Dataset}
We generated a dataset consisting of narrow, corridor-like maps, i.e. maps of width $\ell$ and height $1$.
Each map contains the same set of elements: one player, one box, and one goal. The maps share the same initial configuration in which the goal is positioned at one end of the map, the player at the opposite end, and the box placed in between the two, so that all elements lie along the same row or column.
This choice is motivated by its simplicity: the corridor length $\ell$ is the only map parameter and it serves as a proxy for map difficulty. 
Hence, with just one degree of freedom to account for, we overcome the problem of defining complex measures for solution difficulty: the longer the map, the harder the task.

In our benchmark, we consider map lengths $\ell$ ranging from 5 to 100 in increments of 5. For each map, we generate four augmented variants corresponding to rotations of $90^\circ$, $180^\circ$, and $270^\circ$, as well as the original (unrotated) orientation.
This augmentation strategy reduces the risk of querying the model with data that may have been encountered during pretraining and enables analysis of whether models exhibit orientation-dependent performance.
In total, the evaluation set comprises 80 distinct maps, spanning 20 values of $\ell$ with four orientations each. 
We publicly release our dataset at \url{https://huggingface.co/datasets/Linello/sokobanlevels}.

\subsection{Experimental Setup}
We employ both open and closed weights model, specifically \texttt{DeepSeek R1} \citep{guo2025deepseek}, \texttt{GPT-5} and \texttt{GPT-oss 120B}~\citep{openai2025gpt5,openai2025gptoss120bgptoss20bmodel}.
They are all \textit{reasoning models}, i.e., they are configured to generate an explicit reasoning trace prior to emitting the final answer to the user query.
For GPT models, we don't change the default temperature neither the default reasoning effort (set to medium). 
Instead we cap the maximum number of completion tokens (including both reasoning and final answer tokens) at $32{,}768$.
All inference calls are routed through OpenRouter,\footnote{\href{https://openrouter.ai/}{https://openrouter.ai/}} with the inference provider consistently set to DeepInfra.\footnote{\href{https://deepinfra.com/}{https://deepinfra.com/}}
In light of both computational and financial resource constraints, we limit our empirical analysis to these two primary model families.

\subsubsection{1-shot Inference}
In the first experimental setup, we test the ability of the selected LRMs to solve simple Sokoban puzzles when provided only with the instructions, the mapping of characters as in Table~\ref{tab:ascii-sokoban-notation} and a single demonstration. 
Under this setup, thus, models are by design limited to use exclusively their internal state representations to solve Sokoban puzzles of varying solution lengths. 
The prompts used for all models are described in Appendix \ref{appendix:prompts}.

\subsubsection{LLM-Modulo}
\begin{figure}
\centering
\begin{subfigure}{.6\textwidth}
  \centering
  \includegraphics[width=1\linewidth]{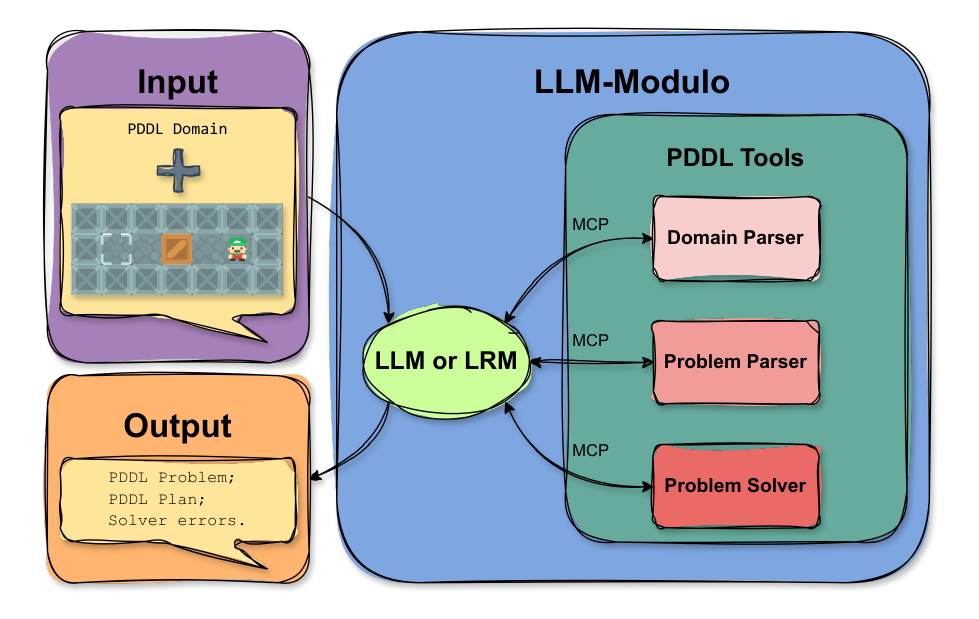}
  \caption{}
  \label{fig:LLM_modulo_diagram}
\end{subfigure}\begin{subfigure}{.4\textwidth}
  \centering
  \includegraphics[width=1\linewidth]{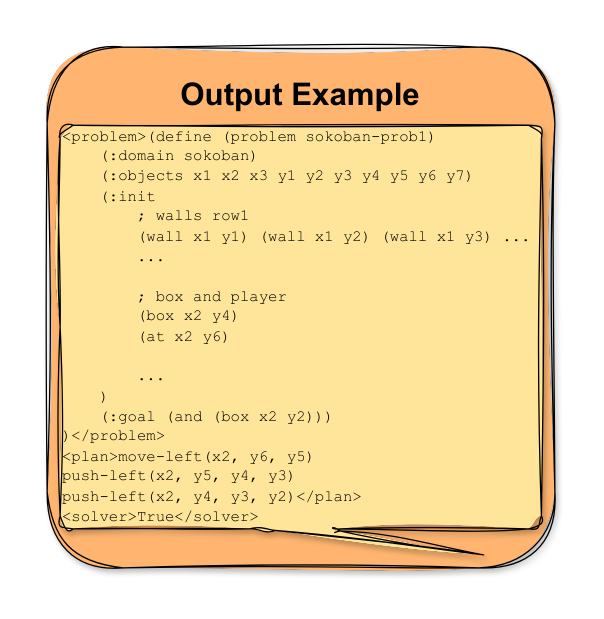}
  \caption{}
  \label{fig:LLM_modulo_input_output}
\end{subfigure}
\caption{Panel (a) represents a simple schema of our LLM-modulo pipeline. The detailed input prompts are collected in Appendix~\ref{sec:agentic_system_prompt}, while an example of output is shown in Panel (b).}
\label{fig:sokoban_LLM_modulo}
\end{figure}

In the second experimental setup, we investigated how Sokoban puzzle–solving performance can be enhanced when LRMs are provided with access to external planning solvers, within an LLM-modulo framework analogous to that of \citet{valmeekam2023planbench}.
To this end, we prompted the models to generate specific instances of planning problems while providing them with a pre-existing, human-authored and verified PDDL domain (Appendix \ref{app:pddl_domain}). 
In this setup, the model is responsible solely for formulating the PDDL problem, which is then processed through an agentic pipeline. 
This workflow utilizes a domain parser to instantiate the formal world representation and a dedicated problem parser that acts as a validator, informing the model whether the generated problem is syntactically and semantically well-formed.
Finally, the pipeline provides access to specialized PDDL planners such as \texttt{Fast-Downward} or \texttt{PyperPlan}, integrated via the \texttt{Unified Planning} library \citep{micheli2025unified,alkhazraji2020pyperplan,helmert2006fast} to solve the problem and get the optimal plan.
All the tools were wrapped and made accessible to the LRMs via a custom Model Context Protocol library \citep{mcpanthropic2024} implemented with \texttt{FastMCP} library \citep{lowin2024fastmcp}.
The design of our architecture is shown in Figure~\ref{fig:sokoban_LLM_modulo}.

The planner tool produces a variety of diagnostic and informational messages that are provided back to the model, including error reports, timing information, and the complete raw response. 
This raw response can be further processed to extract the LURD solution in cases where the problem is successfully solved.
In failure scenarios, the tool returns the encountered errors in natural language to the LRM. 
Errors or warnings are generated in situations such as logically inconsistent or unsatisfiable problems, invalid or inappropriate initial conditions, or when the solver exceeds the maximum allotted execution time (60 seconds). 
The agentic loop ends either with a valid plan or with a message to the final user explaining that, after three failed attempts (which may include having the LRM reformulate the PDDL problem), the agent could not find a satisfactory solution.

The LRM-modulo pipeline is considerably slower than the reasoning-only one. 
It took an average of 75 minutes using \texttt{GPT-5-mini} on an AWS \texttt{t3.xlarge} instance (4 CPUs at 3.1 GHz, 16 GB RAM) to collect the points shown in Figure~\ref{fig:accuracy_LLM_modulo}.
The prompts being used for the experiments are described in Appendix~\ref{appendix:prompts_llm_modulo}.

\subsection{Evaluation}
A solution to a Sokoban instance is defined as a sequence of actions that transforms the system from its initial configuration to the final state, where all boxes are correctly placed on goal positions. 
Multiple valid solutions may exist for the same map, however we restrict the evaluation only to optimal solutions, i.e., sequences that achieve the goal with the minimum possible number of moves. The intrinsic simplicity of our setting makes optimality the natural criterion. 
Clearly, the one-dimensional layout of the map allows only for a unique optimal solution.

\paragraph{Accuracy:}

Given a map of length $\ell$, we define the accuracy in Eq.~\ref{eq:accuracy} as the expectation, over all repetitions and rotations $N$ of the indicator of exact string equality (via Iverson brackets) between the predicted action sequence $\hat{\mathbf{x}}^{(\ell)}$ and the ground-truth sequence $\mathbf{x}^{(\ell)}$, i.e., the fraction of trials in which the two character strings are identical:
\begin{equation}\label{eq:accuracy}
\rm{Accuracy}(\ell) = \frac{1}{N} \sum_{n=1}^{N} \lbrack \hat{\mathbf{x}}^{(\ell)} = \mathbf{x}^{(\ell)} \rbrack.
\end{equation}
Here $N$ is the product of the total number of trials $n_t$ and the number of map rotations $n_r=4$.
Increasing $n_t$ mitigates the intrinsic non-determinism of the obtained solutions, by sampling at multiple seeds.
In the LRM experiments, we set the number of repetitions $n_t$ to 8.
Conversely, in the LRM-modulo experiments, the substantially higher computational and monetary costs imposed stricter constraints. 
We therefore reduced the number of repetitions $n_t$ to 4.

\paragraph{Prefix accuracy:}

Alongside the standard accuracy metric, we define Prefix Accuracy (Eq.~\ref{eq:prefix_accuracy}) to provide a more granular evaluation of model performance. 
This metric calculates the average proportion of correct symbols generated by comparing the predicted and true plans' strings element-wise:

\begin{equation}\label{eq:prefix_accuracy}
\text{PrefixAccuracy}(\ell) = \frac{1}{N} \sum_{n=1}^{N} \frac{[m^{(n)} \leq \ell]}{\ell} \sum_{i=1}^{m^{(n)}} \lbrack \hat{x}_i^{(n)} = x_i^{(n)} \rbrack,
\end{equation}
where $m^{(n)}$ is the length of the predicted plan $\hat{\mathbf{x}}^{(n)}$.
Unlike the hard matching of the standard accuracy metric, prefix-accuracy is more optimistic, rewarding the model for correct partial trajectories even if it stops prematurely. 
However, it remains strictly penalized for overshooting: if the predicted length $m^{(n)}$ exceeds the ground-truth length $\ell$, the score for that trial is $0$. 
For instance, a prediction $\hat{\mathbf{x}}^{(n)} = (\texttt{l, l, l})$ against a ground truth $\mathbf{x}^{(n)} = (\texttt{l, l, l, l})$ yields a score of $3/4$, whereas any prediction exceeding length 4 results in a score of $0$.

\paragraph{Manhattan Distance:}

While string-based metrics evaluate the symbolic fidelity of the action sequence, they do not account for the spatial proximity of the agent to the objective.
We therefore use the Manhattan Distance (Eq.~\ref{eq:manhattan_distance}) to measure the $L_1$ distance between the agent's terminal position and the goal, independent of sequence semantics or environmental obstacles. 

\begin{equation}\label{eq:manhattan_distance}
D(\ell) = \frac{1}{N} \sum_{n=1}^{N} \left( |x^{(n)}_{\text{final}} - x^{(n)}_{\text{goal}}| + |y^{(n)}_{\text{final}} - y^{(n)}_{\text{goal}}| \right)
\end{equation}

Here, $(x^{(n)}_{\text{final}}, y^{(n)}_{\text{final}})$ represents the agent's coordinates after executing all moves in the predicted sequence $\hat{\mathbf{x}}^{(n)}$, starting from the origin $(0,0)$. The goal coordinates $(x^{(n)}_{\text{goal}}, y^{(n)}_{\text{goal}})$ are always at a fixed distance $\ell$ from the origin, specifically $(\pm \ell, 0)$ for $0^\circ/180^\circ$ rotations and $(0, \pm \ell)$ for $90^\circ/270^\circ$ rotations. 

The primary motivation for this metric is to distinguish between ``near-misses'' and total navigational failures.
By measuring spatial displacement, we can quantify whether a model that failed the exact string match nonetheless moved in the correct direction or reached the vicinity of the goal.
This provides a soft failure signal that string-based metrics like Accuracy or Prefix Accuracy cannot capture.

\section{Results}

\subsection{1-shot Inference}
\begin{figure}[htb]
    \centering
    \includegraphics[width=0.98\linewidth]{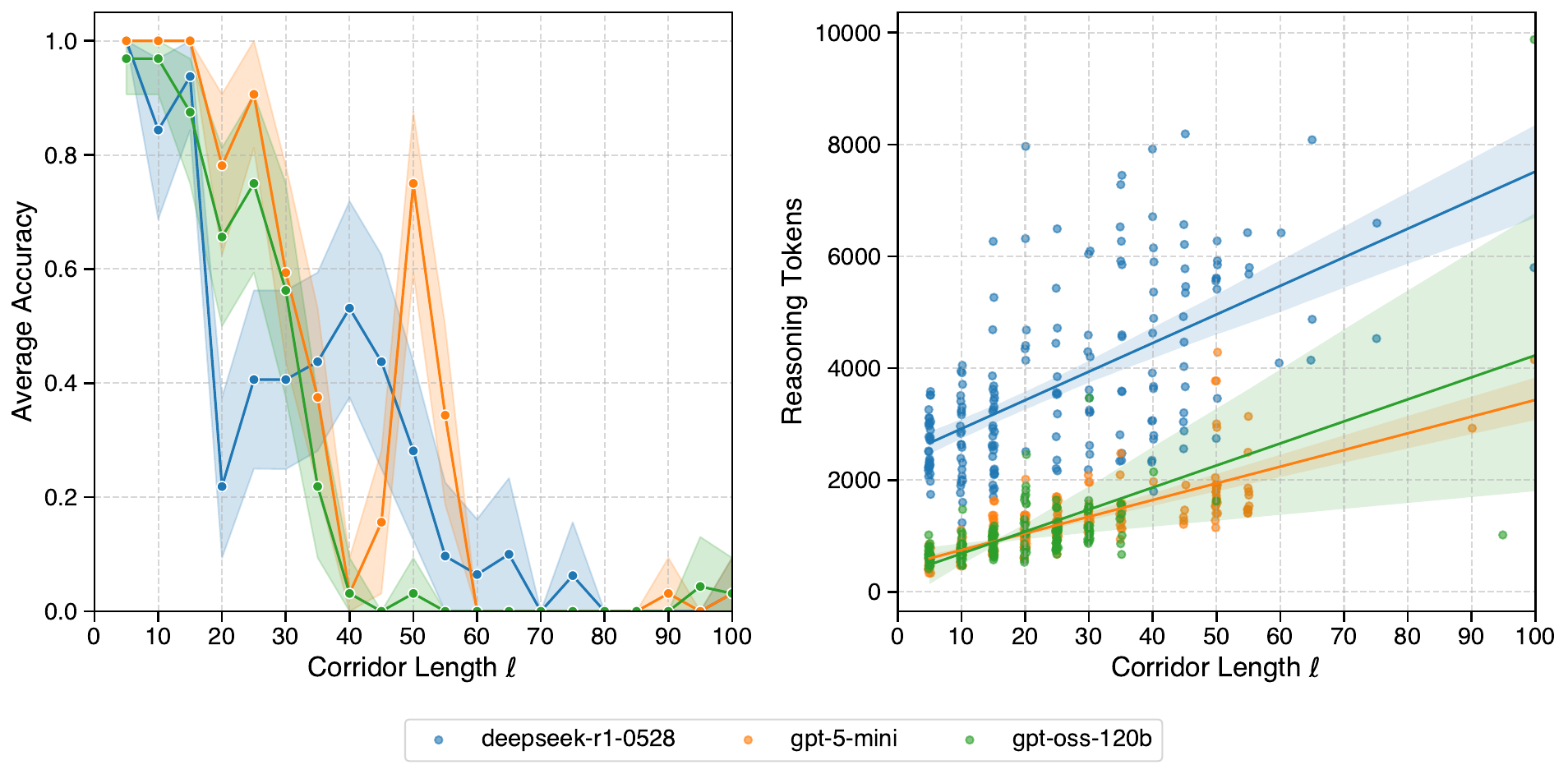}
    \caption{Accuracy and number of reasoning tokens for LRM experiment. Left: average accuracy. Error bars are computed as the 5th and 95th percentile of responses. Right: scaling behaviour of reasoning length against corridor length filtered for correct solutions only.}
    \label{fig:three_plots}
\end{figure}

Figure~\ref{fig:three_plots} summarizes the results in terms of accuracy and total token usage.
The plot on the left of Figure~\ref{fig:three_plots} shows the accuracy as a function of the corridor length, $\ell$, for all tested models.
Similarly to \cite{shojaee2025illusion}, our results show approximately three regions in which the models behave according to different regimes: an easier region where corridor lengths are short, characterized by higher accuracy; an intermediate region characterized by a rapid decrease in accuracy as the length of the corridors increases and a harder region in which the models completely fail to return a correct plan.
These regions are specific to each model. 

Crucially, corridors are \emph{deep but narrow} problems: many sequential steps (depth $d \sim \ell$) with minimal branching. 
In such settings, a small per-step probability $p_w$ of miscounting the size of the map compounds exponentially, yielding success probability $\sim (1-p_w)^\ell$. 
This may explain the three-region performance curve we observe: short corridors tolerate occasional drift, producing a plateau of acceptable accuracy; intermediate lengths mark the onset of exponential decay, while long corridors see near-total collapse as cumulative errors dominate.
We thus believe that the main reason LRMs cannot correctly plan in longer corridors is mainly due to internal counting representation. 
It was indeed shown in \cite{mccoy2024embers} that when asked to count individual characters, LLMs perform better with common characters than uncommon ones (like \texttt{\#}). 
This counting failure can be interpreted through the lens of \cite{lu2025reasoning} ``wandering vs systematic exploration'' framework: maintaining an accurate count over many positions is equivalent to maintaining correct state representations across a chain of transitions. 

As an observation, we report that \texttt{GPT-5-mini} displays an anomalous accuracy peak around $\ell = 50$ which is however hardly explained by the model above.
We believe this effect is likely due to memorization, but without access to internal states models this remains an hypothesis.

Figure~\ref{fig:three_plots} shows the number of output tokens as a function of the corridor length, $\ell$, filtering only for correctly solved Sokoban problems. 
Linear regression analysis reveals that for each model, the number of output tokens for correctly predicted problems increases with the length of the corridor.
We don't observe the counterintuitive scaling mentioned by \cite{shojaee2025illusion} with models declining the request to do very long reasoning to solve complex problems.
Instead, we report the reasoning effort increasing almost linearly with problem complexity, with none of the three models declining our request early.

In the linear regression analysis above, however only a small fraction of the variance is explained due to the noise of the measurements.
This trend is observed only in the region where $\ell < 50$, since for larger corridor's lengths the number of correct predictions decreases significantly for all tested models.
Main parameters of the linear regression fit are collected in Table~\ref{tab:OLS_fit_parameters_LRMs}.

\begin{table}[htb]
    \centering

    \begin{minipage}[t]{0.48\textwidth}
        \centering
        \begin{tabular}{l|c|c}
            \textbf{Model} & \textbf{Slope} & $\mathbf{R^2}$ \\
            \midrule
            \texttt{DeepSeek R1} & $51.1$ & $0.35$ \\
            \texttt{GPT-5-mini} & $29.8$ & $0.62$ \\
            \texttt{GPT-oss 120B} & $39.4$ & $0.40$ \\
        \end{tabular}
        \caption*{\textbf{Correct Answers}}
    \end{minipage}
    \hfill
    \begin{minipage}[t]{0.48\textwidth}
        \centering
        \begin{tabular}{l|c|c}
            \textbf{Model} & \textbf{Slope} & $\mathbf{R^2}$ \\
            \midrule
            \texttt{DeepSeek R1} & 86.3 & 0.25 \\
            \texttt{GPT-5-mini} & 55.2 & 0.14 \\
            \texttt{GPT-oss 120B} & 85.9 & 0.12 \\
        \end{tabular}
        \caption*{\textbf{Wrong Answers}}
    \end{minipage}

    \caption{Fit parameters associated to the linear regressions performed on Figure~\ref{fig:slopes_token_number_LRMs} (see below).}
    \label{tab:OLS_fit_parameters_LRMs}
\end{table}

\begin{figure}[htb]
    \centering
    \includegraphics[width=\linewidth]{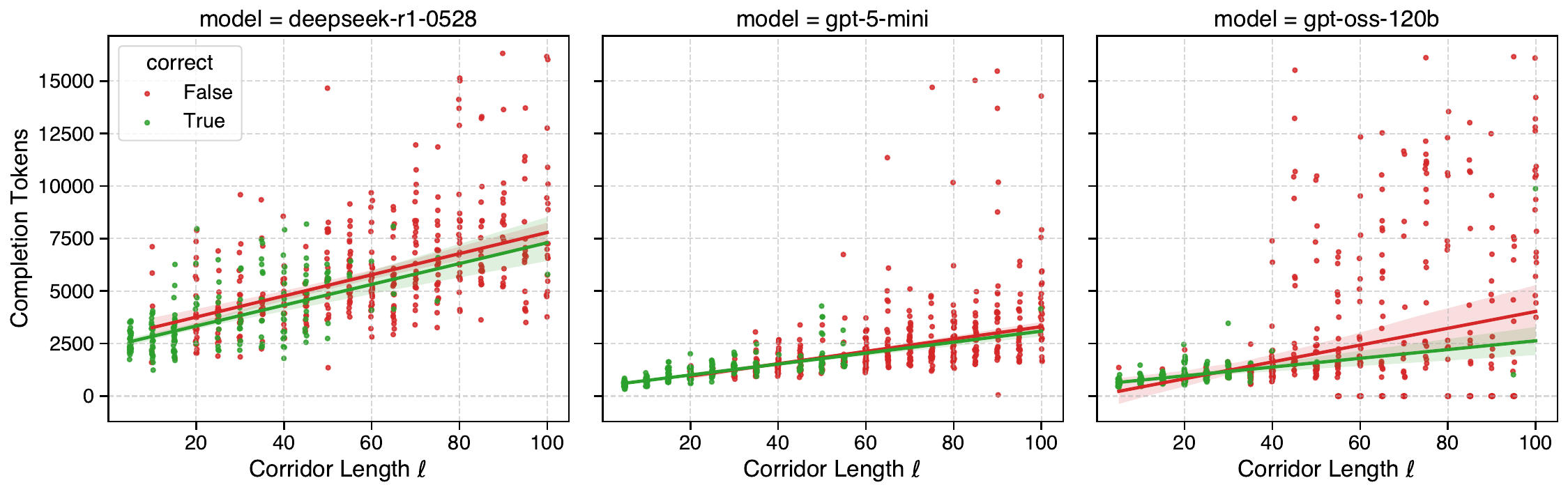}
    \caption{Number of completion tokens produced by each model as a function of corridor length. Separate linear regressions are fitted for correct and incorrect responses, with outliers excluded. A small jitter is added to the x-axis to improve visualization.}
    \label{fig:slopes_token_number_LRMs}
\end{figure}

In Figure~\ref{fig:slopes_token_number_LRMs} we further analyze the number of emitted tokens as a function of the corridor length parameter $\ell$, considering both correct and incorrect answers. 
Unlike \cite{shojaee2025illusion} which observed a counterintuitive reduction in the reasoning effort for problems above a certain threshold of difficulty, we observe a steady increase in the number of output tokens.
What we found shows that the difficulty of a problem is not characterized by the decrease in the reasoning effort, but instead by the substantially higher variability in token counts of incorrect answers compared to correct ones.
This suggests that when the model diverges from the correct reasoning trajectory, it can fail in multiple ways, whereas successful completions remain more concise and consistent, likely an effect of inductive bias of Group Reinforcement Policy Optimization (GRPO) post-training, where concise reasoning traces are preferred to lengthy ones \citep{sui2025stop}.
To quantify this effect, we fit a robust regression of completion tokens against corridor's length for each model.
Both slope and intercept appear model-specific: more efficient models, such as \texttt{GPT-5-mini}, show lower slopes and reduced variability across both correct and incorrect responses.
Another relevant distinction can be made, highlighting differences in models' calibration.
\texttt{DeepSeek-R1} and \texttt{GPT-5-mini} display similar slopes and intercepts between correct and incorrect predictions, \texttt{GPT-oss-120B} instead reflect large differences in the regression parameters.
A recurrent behavior is that for longer corridors, LRMs often reach the maximum allowed number of output tokens.
After a qualitative inspection of the reasoning traces, we observed that the main reason this happens is that the models get stuck in repeating the same action or reasoning frame over and over until they reach the token limit.
We report the reasoning traces for the interested user at \url{https://anonymous.4open.science/r/sokoban_traces/}

This repetitive looping behavior exemplifies what Lu et al. \citep{lu2025reasoning} classify as \emph{unnecessary exploration} and failure to maintain a visited-state set. In a systematic search, an agent would track which configurations (or reasoning states) have already been explored and avoid revisiting them. The token-limit exhaustion we observe suggests that LRMs lack such memory: they repeatedly propose the same moves or reasoning steps without recognizing the cycle. 
This is evidence of \emph{wandering} rather than systematic planning: the model explores aimlessly rather than pruning redundant paths. 
In a corridor setting, where the state space is essentially linear, even a simple mental tape of visited positions would suffice to prevent loops; the inability to maintain it indicates a fundamental deficit in structured state tracking.

\begin{figure}[htb]
\centering
\begin{subfigure}{.5\textwidth}
  \centering
  \includegraphics[width=0.9\linewidth]{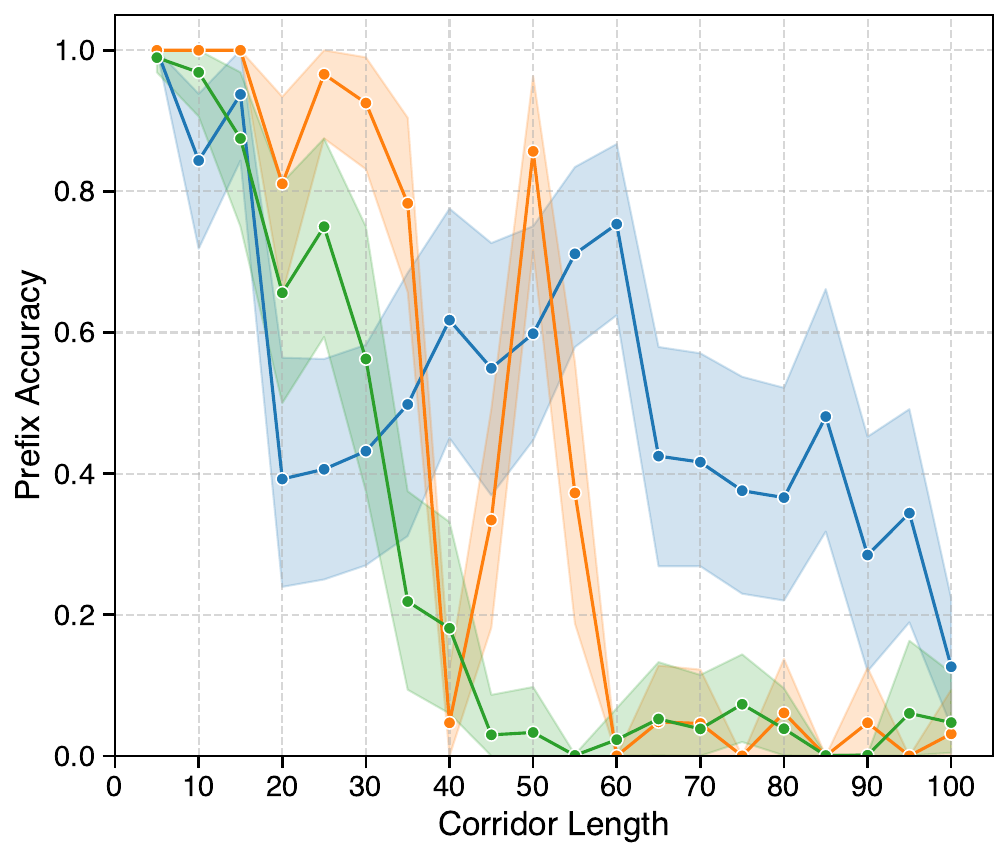}
  \caption{}
  \label{fig:prefix_accuracy_LRMs}
\end{subfigure}\begin{subfigure}{.5\textwidth}
  \centering
  \includegraphics[width=0.9\linewidth]{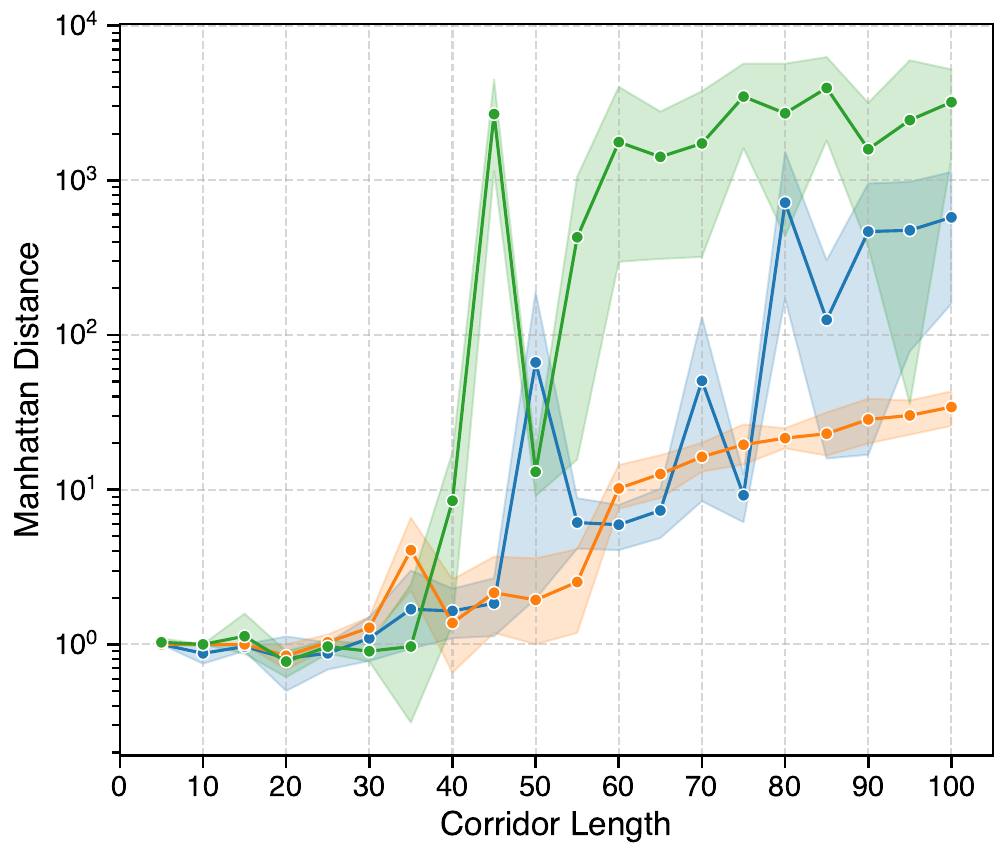}
  \caption{}
  \label{fig:manhattan_distance_LRMs}
\end{subfigure}
\caption{Other useful metrics represented as functions of the corridor's length. Panel (a) represents prefix accuracy, computed as described in Equation~\ref{eq:prefix_accuracy}. Panel (b) represents Manhattan distance, computed as in Equation~\ref{eq:manhattan_distance}. Models' colors are the same as in Figure~\ref{fig:three_plots}.}
\label{fig:prefix_and_manhattan_LRMs}
\end{figure}

In Figure~\ref{fig:prefix_and_manhattan_LRMs} we analyze our data from the point of view of prefix accuracy and  Manhattan distance.
The metrics show a decreasing trend for all models that is similar to that represented in Figure~\ref{fig:three_plots}.
Some patterns, like the peak at $\ell=50$ for \texttt{GPT-5-mini} and the increase in accuracy around the central region for \texttt{DeepSeek R1}, are further accentuated. 
This highlights that the main source of errors in most Sokoban problems is related to counting mistakes.
In terms of Manhattan distance, the optimal solution would have distance one as the player and the goal are separated by the box.
However, as observed sometimes the player is positioned exactly on the goal, thus ignoring the spatial constraints of the problem.

These violations, where the predicted sequence places the player on the goal position despite walls and box, are instances of what \cite{lu2025reasoning} terms \emph{invalid exploration}. 
In a valid state-transition graph, certain moves (e.g., walking through walls, teleporting over boxes) are inadmissible. 
When a model proposes such transitions, it demonstrates that its internal representation does not faithfully track the game's physics and its constraints. 
LLMs hallucinate states unreachable under the true transition rules, producing reasoning traces that are syntactically plausible but structurally incoherent for the problem.
The fact that even advanced reasoning models exhibit these errors underscores a core limitation: without explicit state-transition verification, test-time scaling cannot guarantee adherence to problem constraints and rules.

\subsection{LLM-Modulo}
\begin{figure}[htb]
\centering
\begin{subfigure}{.5\textwidth}
  \centering
  \includegraphics[width=0.9\linewidth]{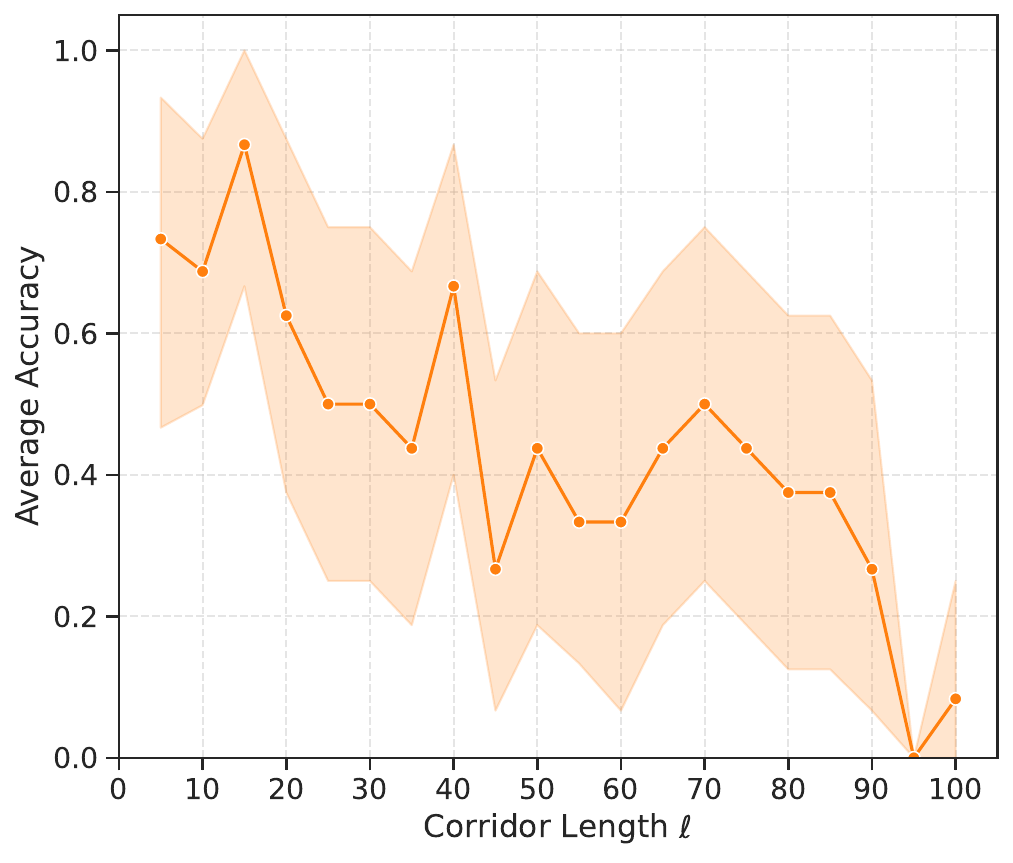}
  \caption{}
  \label{fig:accuracy_LLM_modulo}
\end{subfigure}\begin{subfigure}{.5\textwidth}
  \centering
  \includegraphics[width=0.9\linewidth]{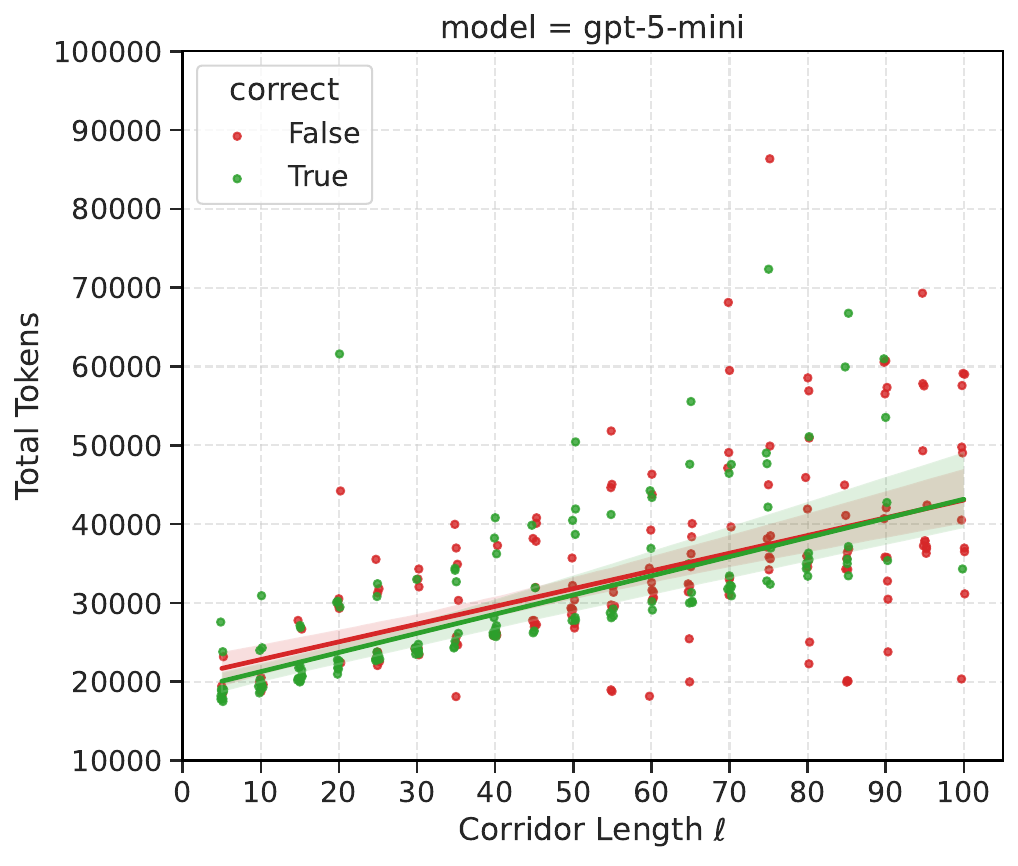}
  \caption{}
  \label{fig:tokens_LLM_modulo}
\end{subfigure}
\caption{Results of the LLM-Modulo approach for \texttt{GPT-5-mini}. Panel (a) represents the average accuracy (Eq.~\ref{eq:accuracy}). Panel (b) shows the counts of total tokens, for correct (green) and incorrect (red) predictions, together with separate robust linear regressions. Error ribbons are computed as 5 and 95 percentiles. A small jitter is added to the x-axis to improve visualization.}
\label{fig:acc_tokens_LLM_modulo}
\end{figure}

Figure~\ref{fig:acc_tokens_LLM_modulo} shows the main results of the LRM-Modulo approach based on classical PDDL planning tools (domain, problem parsers and a problem solver). 
Unfortunately, preliminary experiments showed that not many models are both affordable in terms of costs and effectiveness in tool-use tasks.
Typical failures we encountered in testing models like \texttt{DeepSeek R1}, \texttt{Gemini-2.5-Flash-Preview}, and \texttt{Claude-3.5-Haiku} include: limited capability to interact with tools, difficulty to generate coherent PDDL problems even for simpler Sokoban problems, and inability to stop calling tools after a given number of attempts.
\texttt{GPT-5-mini} resulted as the only model among the tested ones that could generate accurate PDDL problems and interact with tools while following prompt instructions. 
Due to the higher costs of the experiments in the LMR-Modulo setting we limited the experiment's repetitions per corridor rotation $n_t$ to four. 
Nonetheless, \texttt{GPT-5-mini} exhibits high stability in the accuracy and in the number of reasoning tokens, allowing to maintain a valid evaluation even with a lower number of repetitions.

\begin{figure}
\centering
\begin{subfigure}{.5\textwidth}
  \centering
  \includegraphics[width=0.9\linewidth]{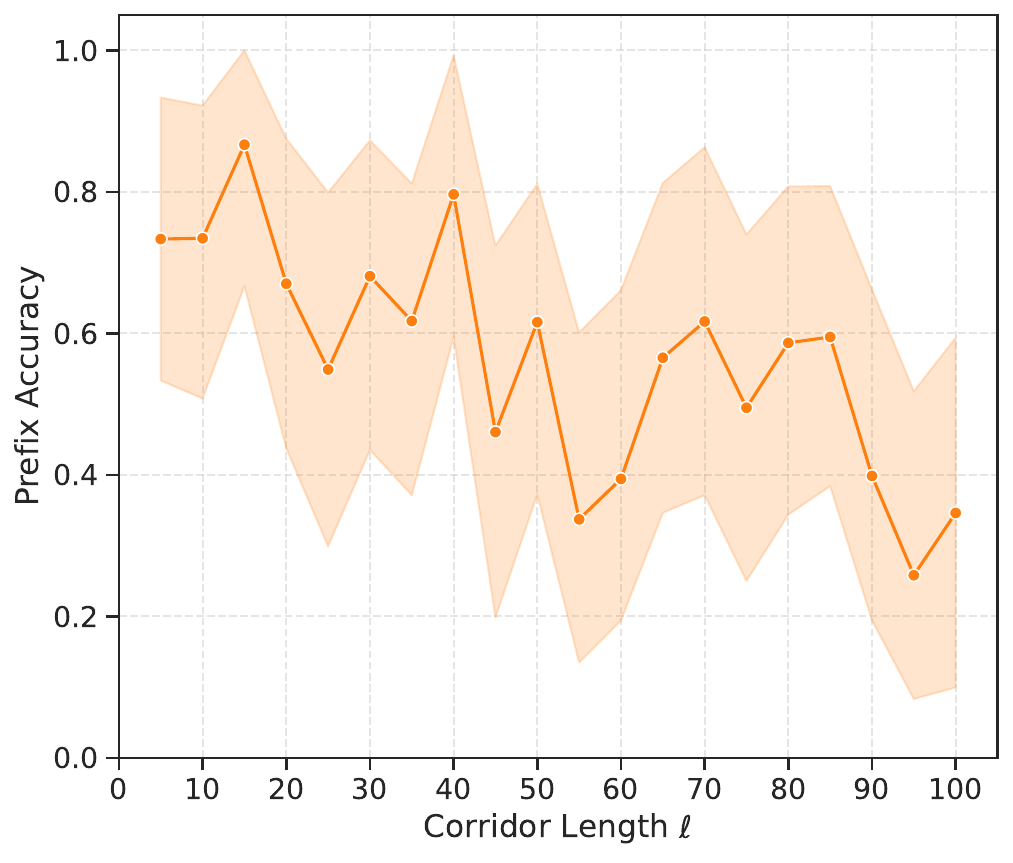}
  \caption{}
  \label{fig:prefix_accuracy_LLM_modulo}
\end{subfigure}\begin{subfigure}{.5\textwidth}
  \centering
  \includegraphics[width=0.9\linewidth]{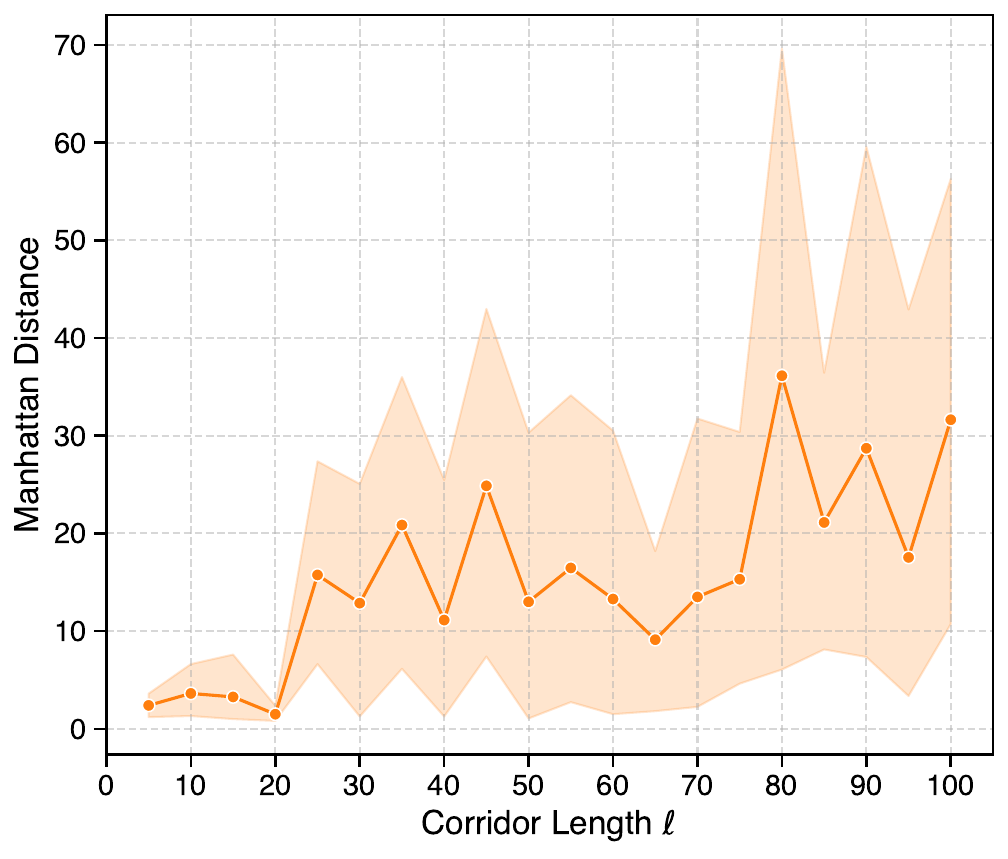}
  \caption{}
  \label{fig:manhattan_distance_LLM_modulo}
\end{subfigure}
\caption{Additional metrics for \texttt{GPT-5-mini} in the LLM-modulo framework. Panel (a) represents prefix accuracy (Eq.~\ref{eq:prefix_accuracy}). Panel (b) shows Manhattan distance (Eq.~\ref{eq:manhattan_distance}).}
\label{fig:prefix_accuracy_manhattan_distance_LLM_modulo}
\end{figure}

In Figure~\ref{fig:accuracy_LLM_modulo}, the absence of sharp peaks and the slower descending trend highlights a more regular accuracy behavior compared to that shown in Figure~\ref{fig:three_plots} for LRMs alone.
However, a higher variability is observed and the main reason is due to non-homogeneous performances across experimental trials and map rotations for a fixed corridor length.
Visual inspection of the results reveals a significant imbalance between accuracy in vertical and horizontal corridors (Figure~\ref{fig:accuracies_rotations_LLM_modulo}, Appendix~\ref{sec:llm_modulo_map_rotations}) showing that, also in LRM-modulo setting, models struggle to solve vertical corridors.
At the same time, a detailed analysis of the source of these errors indicates two main causes of failure.
One occurs when there are syntax errors in the generated PDDL problems, producing error messages when calling the solver tool.
The other occurs when generated PDDL problems are syntactically correct but do not represent the actual Sokoban problem.
In our data, first-type errors just occur 7 times out of all four trials of the 80 corridor configurations, meaning that in the large majority of cases the solver tool compiles correctly and produces a valid solution.
The charts depicting the prefix accuracy and the Manhattan distance, represented in Figure~\ref{fig:prefix_accuracy_manhattan_distance_LLM_modulo}, confirm that in many cases the generated PDDL representation of the Sokoban problems leads to solutions in which the player, although moving in the right direction, does not reach the number of moves required to push the box towards the goal position.  

By utilizing an expert-validated PDDL domain and a solver strictly governed by logical constraints, we have effectively eliminated the risk of invalid transitions. 
Hence, the primary challenge for these models lies in maintaining a consistent internal representation of the spatial environment. 
Evidence suggests this difficulty may stem from a fundamental limitation in the models' ability to precisely quantify the dimensions of the map.


 \section{Conclusions}
The assessment of the long-horizon planning capacities of language models is both required and attainable. 
Adhering to the principle of beginning with simplistic settings before advancing to more intricate ones, we propose utilizing a simplified version of Sokoban as a controlled environment to evaluate planning capabilities. 
Our observations, in agreement with prior research, suggest that long planning abilities of LLMs may not only be related to problem complexity but from lack of more elementary initial abilities like counting.

We observe that even advanced reasoning models struggle to solve Sokoban instances that require anticipating the goal state more than 25–30 moves ahead.
We discussed several possible causes for this limited performance in the limitations section, including the absence of textual cues and the inability to reliably store intermediate states within model hidden representations.

Equipping language models with a PDDL parser, validator, and solver slightly improves planning capabilities on average, but not enough to overcome the lack of inherent spatial grounding. 
We found that the basic, initial inability to track counts remains a persistent bottleneck. 
This issue surfaces even in LLM modulo settings where external symbolic engines are used, proving that offloading logic to a solver cannot fully fix a model that cannot faithfully represent space and constraints.

More broadly, our observations align with recent characterizations of reasoning models as ``wanderers'' rather than systematic explorers: linear corridors exemplify a setting where minimal branching but substantial depth exposes how small per-step errors in state tracking (counting drift, visited-state amnesia, invalid transitions) compound exponentially. 
Consequently, test-time scaling alone cannot overcome these structural limitations without architectural innovations, short horizon error tracking or explicit symbolic grounding.

\subsection{Current limitations and future work}
Our study is intentionally narrow; here we outline the main constraints and threats to validity.
We focus on one-box linear corridors, which test long-horizon counting and state maintenance rather than the full difficulty of multi-box Sokoban with deadlocks.
Thus, the benchmark provides only a lower bound on planning ability.
For evaluation, we use exact-plan validation against a reference generator.
Although this is stricter than necessary in general Sokoban, where multiple optimal plans may exist, it is suitable for corridors; future work will instead use solver-based verification to handle maps with multiple valid solutions.
We also find sensitivity to prompt formatting, especially orientation-related effects such as the many newlines in vertical maps.
Alternative encodings, such as row/column numbered grids or other textual cues, may reduce this issue.
Another variability source is model metadata and provider backends: although all calls go through one routing layer, backend implementations and model revisions can change over time.
We log identifiers and dates, but some instability is inherent in API-based evaluations.
Pretraining contamination is another concern; corridor rotations lower the chance that specific plans were memorized but do not eliminate it.
Finally, corridor tasks have limited external validity, since success or failure may not transfer to richer planning domains.
We treat these settings mainly as a sanity check, with follow-up experiments planned to add obstacles, branching structures, and deadlocks.

\section*{Societal Impact}
This paper presents work whose goal is to advance the field of Machine Learning through clearer diagnostics of long-horizon planning.
While any benchmark could have indirect downstream effects by steering research agendas, we do not identify specific societal risks unique to this work beyond standard concerns about evaluation misuse.
We therefore do not highlight any particular societal impacts at this time.

 
\bibliographystyle{tmlr}
\bibliography{references}

\appendix
\section{Prompts for 1-shot Inference Settings}\label{appendix:prompts}
In this section we report the detailed prompts that we have used throughout our experiments with reasoning models alone.
Prompts are direct, no Chain of Thought elicited as it is known that it may hamper internal reasoning on LRMs.
A simple solved problem is provided as the 1-shot example.

\begin{tcolorbox}[
    colback=blue!5, colframe=blue!80!black, coltitle=white, title=System Prompt, fonttitle=\bfseries,  fontupper=\ttfamily,
    boxrule=0.8pt,  arc=1mm, left=6pt, right=6pt, top=6pt, bottom=6pt]
\begin{lstlisting}[basicstyle=\ttfamily\tiny,breaklines=true]
You are an assistant that helps in solving assigned Sokoban games.
Your task is to examine the provided Sokoban problem and find a solution.
All provided Sokoban problems are assigned in form of ASCII maps.
The mapping is the following:

```
    @ - Player
    + - Player on Goal
    $ - Box
    * - Box on Goal
    . - Goal (Empty)
    # - Wall Brick
```

The game is solved when the box is pushed into the goal position, hence when the position of `$` coincides with the position of `.`.
The player can move in all the empty spaces of the ASCII map while respecting the walls.
When the player is adjacent to a box, the player can push the box into an adjacent empty space. 
After pushing a box, the new position of the agent will be the position of the box before the push.
The player cannot pull the box, only push it.
The actions you can perform in the game are:

```
    L - Move Left
    R - Move Right
    U - Move Up
    D - Move Down
```

All provided problems CAN be solved.
You must give your solution in form of a sequence of allowed actions, separated by commas.
You must give only the sequence of actions, without any additional text or explanation.
You must enclose your solution inside the tags <plan> </plan>.

The following is an example of a Sokoban problem and its solution:

Problem:

#####
#@  ##
## $ ##
 #    #
 ##. ##
  ## #
   ###

Solution:

<plan>
R,R,D,D
</plan>
\end{lstlisting}
\end{tcolorbox}

\begin{tcolorbox}[
    colback=red!5, colframe=red!80!black, coltitle=white,       title=User Prompt, fonttitle=\bfseries,  fontupper=\ttfamily,
    boxrule=0.8pt,        arc=1mm,              left=6pt, right=6pt, top=6pt, bottom=6pt, ]
\begin{lstlisting}[basicstyle=\ttfamily\tiny,breaklines=true]
Here is the Sokoban problem to solve, enclosed in triple backtics:

```
{{ sokoban_map }}
```
\end{lstlisting}
\end{tcolorbox}

\section{Prompts for LLM-Modulo Settings}\label{appendix:prompts_llm_modulo}
In this section, we show the system prompt we used for the experiments on LLM-Modulo settings. 
The user prompt remains the same as shown in Appendix~\ref{appendix:prompts}. The system prompt includes the human-designed PDDL domain of a typical Sokoban game (\href{https://verificationglasses.wordpress.com/2021/01/02/sokoban-pddl}{https://verificationglasses.wordpress.com/2021/01/02/sokoban-pddl}).

\subsection{System Prompt}
Here are the system prompts and the PDDL domain being used for the experiments in LLM-Modulo settings. 
The model is just required to generate the PDDL problem to be sent to the solver.
\label{sec:agentic_system_prompt}

\begin{tcolorbox}[
    colback=blue!5,       colframe=blue!80!black, coltitle=white,       title=System Prompt, fonttitle=\bfseries,  fontupper=\ttfamily,
    boxrule=0.8pt,        arc=1mm,              left=6pt, right=6pt, top=6pt, bottom=6pt, ]
\begin{lstlisting}[basicstyle=\ttfamily\tiny,breaklines=true]
You are an assistant that helps in solving assigned Sokoban games.
All provided Sokoban problems are assigned in form of ASCII maps and CAN be solved.
The mapping is the following:

\`\`\`
    @ --- Player
    + --- Player on Goal
    \$ --- Box
    * --- Box on Goal
    . --- Goal (Empty)
    \# --- Wall Brick
\`\`\`

Given the PDDL domain of a generic Sokoban game, your task is to generate a valid PDDL problem representation of the provided ASCII Sokoban problem.
Once you generate the PDDL problem, your final goal is to find a plan that solves the problem.
You have access to a set of tools to help you achieve your goal. 
Always use the solve\_problem tool to solve the problem, do not try to solve it yourself.
If you encur in any error while solving a problem with the tool, try to fix it and call the tool again.
Retry up to 3 times at maximum if needed.

Here is the PDDL Sokoban domain, enclosed in triple backtics:
\`\`\`
\{\{PDDL\_domain\}\}
\`\`\`

IMPORTANT:
Your final answer must contain both the the PDDL problem and the solution to the problem without any additional text or explanation.
You must separately enclose the PDDL problem inside the tags <problem> </problem>, and the solution inside the tags <plan> </plan>.
If, after the third attempt, you are unable to get a solution from the solver, provide the error message you received from the tool inside the <plan> </plan> tags.
If at the end of your process the solve\_problem tool gets called without errors and returns a solution, write <solver>True</solver>, otherwise write <solver>False</solver>.

Example output:
\`\`\`yaml
problem: <problem>PDDL problem here</problem>
plan: <plan>PDDL plan from solver here</plan>
solver: <solver>Boolean checking whether solve\_problem tool was called successfully</solver>
\`\`\`
\end{lstlisting}
\end{tcolorbox}

\subsection{PDDL Domain}\label{app:pddl_domain}
Here the human authored PDDL domain used in the above system prompt is reported for completeness.
\begin{tcolorbox}[
    colback=gray!5,       colframe=gray!80!black, coltitle=white,       title=PDDL Domain, fonttitle=\bfseries,  fontupper=\ttfamily,
    boxrule=0.8pt,        arc=1mm,              left=6pt, right=6pt, top=6pt, bottom=6pt, ]
\begin{lstlisting}[basicstyle=\ttfamily\tiny,breaklines=true]
(define (domain sokoban)
    (:predicates (wall ?x ?y) (box ?x ?y) (at ?x ?y) (inc ?p ?pp) (dec ?pp ?p))
    (:action move-up
        :parameters (?x ?y ?xn)
        :precondition (and (at ?x ?y) (not (wall ?xn ?y)) (not (box ?xn ?y)) (dec ?x ?xn))
        :effect (and (not (at ?x ?y)) (at ?xn ?y))
    )
    (:action move-down
        :parameters (?x ?y ?xn)
        :precondition (and (at ?x ?y) (not (wall ?xn ?y)) (not (box ?xn ?y)) (inc ?x ?xn))
        :effect (and (not (at ?x ?y)) (at ?xn ?y))
    )
    (:action move-right
        :parameters (?x ?y ?yn)
        :precondition (and (at ?x ?y) (not (wall ?x ?yn)) (not (box ?x ?yn)) (inc ?y ?yn))
        :effect (and (not (at ?x ?y)) (at ?x ?yn))
    )
    (:action move-left
        :parameters (?x ?y ?yn)
        :precondition (and (at ?x ?y) (not (wall ?x ?yn)) (not (box ?x ?yn)) (dec ?y ?yn))
        :effect (and (not (at ?x ?y)) (at ?x ?yn))
    )
    (:action push-up
        :parameters (?x ?y ?xn ?xnn)
        :precondition (and (at ?x ?y) (not (wall ?xn ?y)) (box ?xn ?y) (dec ?x ?xn) (not (wall ?xnn ?y)) (not (box ?xnn ?y)) (dec ?xn ?xnn))
        :effect (and (not (at ?x ?y)) (at ?xn ?y) (not (box ?xn ?y)) (box ?xnn ?y))
    )
    (:action push-down
        :parameters (?x ?y ?xn ?xnn)
        :precondition (and (at ?x ?y) (not (wall ?xn ?y)) (box ?xn ?y) (inc ?x ?xn) (not (wall ?xnn ?y)) (not (box ?xnn ?y)) (inc ?xn ?xnn))
        :effect (and (not (at ?x ?y)) (at ?xn ?y) (not (box ?xn ?y)) (box ?xnn ?y))
    )
    (:action push-right
        :parameters (?x ?y ?yn ?ynn)
        :precondition (and (at ?x ?y) (not (wall ?x ?yn)) (box ?x ?yn) (inc ?y ?yn) (not (wall ?x ?ynn)) (not (box ?x ?ynn)) (inc ?yn ?ynn))
        :effect (and (not (at ?x ?y)) (at ?x ?yn) (not (box ?x ?yn)) (box ?x ?ynn))
    )
    (:action push-left
        :parameters (?x ?y ?yn ?ynn)
        :precondition (and (at ?x ?y) (not (wall ?x ?yn)) (box ?x ?yn) (dec ?y ?yn) (not (wall ?x ?ynn)) (not (box ?x ?ynn)) (dec ?yn ?ynn))
        :effect (and (not (at ?x ?y)) (at ?x ?yn) (not (box ?x ?yn)) (box ?x ?ynn))
    )
)
\end{lstlisting}
\end{tcolorbox}

\section{LLM-Modulo: Map Rotations}
\label{sec:llm_modulo_map_rotations}
In this section we show the results of the LLM-modulo setting in all map rotations separately. Accuracies are just averaged over the four experiment trials.

\begin{figure}[htb]
    \centering
    \includegraphics[width=1\linewidth]{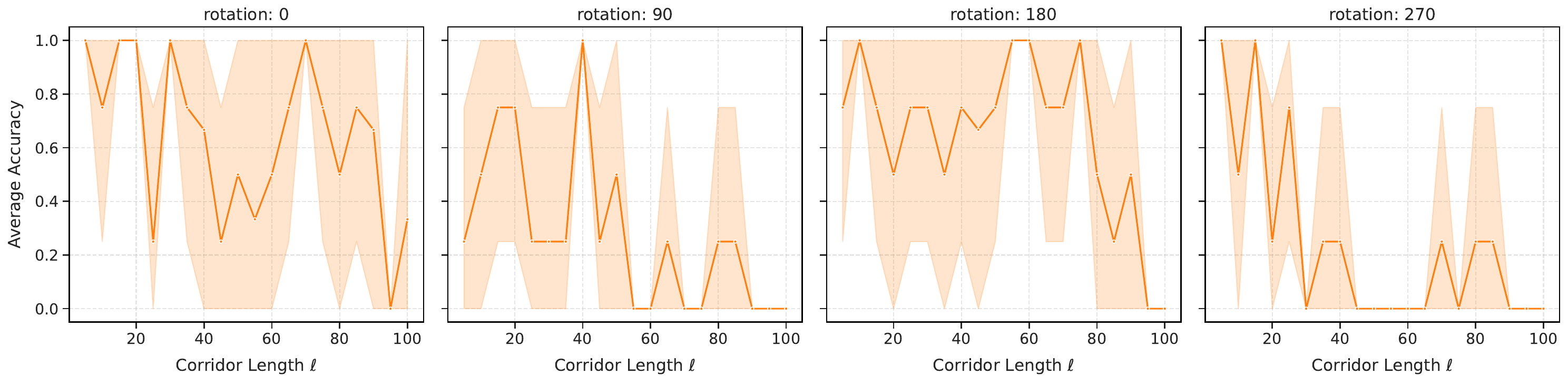}
    \caption{GPT-5 mini accuracies in LLM-modulo setting, averaged over four experiment trials on each Sokoban corridor rotation.}
    \label{fig:accuracies_rotations_LLM_modulo}
\end{figure}
 
\end{document}

%% file: math_commands.tex
\usepackage{amsmath,amsfonts,bm}

\def\eqref#1{equation~\ref{#1}}

\def\1{\bm{1}}

\DeclareMathAlphabet{\mathsfit}{\encodingdefault}{\sfdefault}{m}{sl}
\SetMathAlphabet{\mathsfit}{bold}{\encodingdefault}{\sfdefault}{bx}{n}

